\pdfoutput=1

\documentclass[11pt]{article}

\usepackage{ACL2023}

\usepackage{times}
\usepackage{latexsym}

\usepackage[T1]{fontenc}

\usepackage[utf8]{inputenc}

\usepackage{microtype}

\usepackage{inconsolata}

\usepackage{amsmath}
\usepackage{amssymb}

\usepackage{microtype}
\usepackage{graphicx}
\usepackage{booktabs}
\usepackage{makecell}
\usepackage{multirow}

\usepackage{svg}

\usepackage{color}

\usepackage{CJKutf8}

\title{Understanding Client Reactions in Online Mental Health Counseling}

\author{\textbf{Anqi Li}$^{1,2}$\thanks{\ \ Equal Contribution.}, \textbf{Lizhi Ma}$^{2}$\footnotemark[1], \textbf{Yaling Mei}$^{2}$ \\ \textbf{Hongliang He}$^{1,2}$, \textbf{Shuai Zhang}$^{1,2}$, \textbf{Huachuan Qiu}$^{1,2}$, \textbf{Zhenzhong Lan}$^{2}$\thanks{\ \ Corresponding Author.} \\
$^{1}$ Zhejiang University \\ 
$^{2}$ School of Engineering, Westlake University \\
\texttt{\{lianqi, malizhi, lanzhenzhong\}@westlake.edu.cn} \\
}

\begin{document}
\maketitle
\begin{abstract}

Communication success relies heavily on reading participants' reactions. Such feedback is especially important for mental health counselors, who must carefully consider the client's progress and adjust their approach accordingly. However, previous NLP research on counseling has mainly focused on studying counselors' intervention strategies rather than their clients' reactions to the intervention. This work aims to fill this gap by developing a theoretically grounded annotation framework that encompasses counselors' strategies and client reaction behaviors. The framework has been tested against a large-scale, high-quality text-based counseling dataset we collected over the past two years from an online welfare counseling platform. Our study show how clients react to counselors' strategies, how such reactions affect the final counseling outcomes, and how counselors can adjust their strategies in response to these reactions. We also demonstrate that this study can help counselors automatically predict their clients' states~\footnote{\ You can access our annotation framework, dataset and codes from https://github.com/dll-wu/Client-Reactions.}.

\end{abstract}


\section{Introduction}

There can be no human relations without communication, yet the road to successful communication is paved with obstacles~\citep{luhmann1981improbability}. Given the individuality and separateness of human consciousness, it is hard to guarantee one can receive the message sent by another. Even if the message is fully understood, there can be no assurance of its acceptance. By getting feedback from their partners, communicators can better understand their communicative states. This allows communicators to adjust their communication strategies to fit better their communication environment, which is crucial for successful communication. However, most work on improving the success rates in communication, such as persuasion~\citep{wang2019persuasion} and mental health support~\citep{zhang2019finding,2020Balancing,liu2021towards}, focuses on speakers' strategies. But little research is on how listeners' reactions shape trajectories and outcomes of conversations. In this work, we address the gap by
examining how to use the reactions of clients to predict and improve the quality of psychological counseling, a field that has profound societal and research impact.

\begin{figure}[t!]
    \centering
    \includegraphics[scale=0.21]{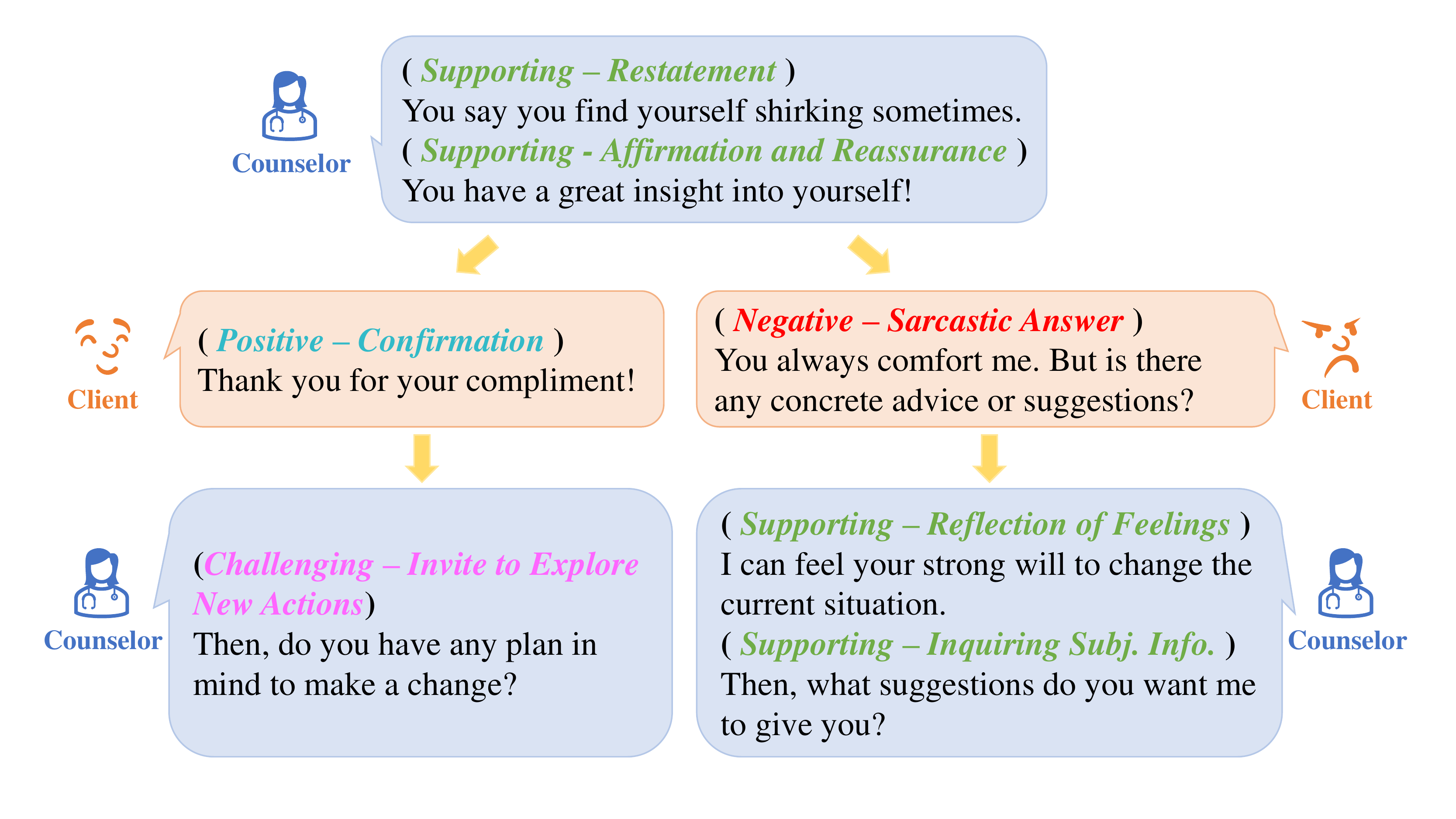}
    \caption{Examples of how counselors adjust their strategies according to their client's reactions. }
    \label{fig:dialog}
\end{figure}

Psychological counseling is one of the most challenging and skillful forms of communication~\citep{althoff2016large}. Counselors take clients through their mental health concerns while balancing the stress they are experiencing~\citep{2020Balancing}. 
To do it well, counselors rely on training and on continuing experience with clients to acquire the consultative skills. However, it is difficult for counselors to get direct feedback on their interventions from clients in practice~\citep{zhang2019finding}. Besides, due to the lack of accurate assessments of general counseling interventions~\citep{tracey2014expertise}, the prior studies found no noticeable improvement or effectiveness of counselors' interventions after training or counselings ~\citep{dawes2009house,hill2015training,goldberg2016psychotherapists}. As a result, some have even argued that psychological counseling is"\textit{ a profession without any expertise}"~\citep{shanteau1992competence}. In this regard, one solution to facilitate counselors noticing the effectiveness of interventions is to know clients' feedback during counseling conversations.

However, researchers in the field mainly study counselors' skills and language patterns to provide feedback on interventions~\citep{althoff2016large,zhang2019finding,perez2019makes}. They first separate counselings into two groups, high-quality and low-quality counselings. Then, features of counselors' interventions, such as language diversity, ability to handle ambiguity and make progress, are analyzed. In the end, the general patterns of the features of good counseling are reported. Nonetheless, apart from the counselors' interventions, the counseling, as a process of interactive communication, also includes clients' reactions~\citep{avdi2007discourse}. Importantly, the clients' reactions towards counselors' intervention reflect the feedback on the effectiveness of the interventions~\citep{ribeiro2013TCCS}. Thus, to complete the assessment of counselors' interventions from the client's perspective and to provide feedback for counselors, we are motivated to categorize the clients' reactions although identifying their reactions in the psychological counseling is difficult, even more so than categorizing counselors' interventions~\citep{lee2019identifying, sharma2020computational-empathy}. 

In this paper, we introduce a theoretically grounded annotation framework to map each turn of the conversation into counselors' intentions and their clients' reactions. The framework is applied to label a large-scale text-based Chinese counseling dataset collected from an online welfare counseling platform over the last two years.  

Using the annotation, we analyze the associations between clients' reactions and behaviors in the counselling conversation and their assessment of conversation effectiveness. We demonstrate that the counselors’ different intentions and strategies elicit different follow-up reactions and behaviors from the clients. Following this analysis, we examine how counselors should adjust their strategies to encourage clients' positive behaviors based on different conversation stages and historical interaction patterns. We also analyze how the counselors address the clients' behaviors that negatively impact the conversation effectiveness. Along with the automatic annotation classifiers we built, the findings of above analyses would help develop user-centered mental health support dialog systems.

\section{Related Work}

We mostly draw inspiration from conversational analysis in NLP and psychotherapy.

Despite the abundance of NLP research relating to emotional chat~\citep{zhou2018emotional}, emotional support~\citep{liu2021towards}, and psycho-counseling~\citep{althoff2016large}, in most cases, these studies are still in their infancy. Human-human interaction patterns are rarely studied due to the lack of large-scale conversational datasets~\citep{huang2020challenges}. Meanwhile, the main research focus is either on proposing new datasets or studying consultation skills.

\noindent{\textbf{Dataset for Mental Health Support.}} Because of the sensitive nature of mental health data, most of the available mental health support conversation corpora are collected from public general social networking sites or crowdsourcing~\citep{sharma2020computational-empathy, Harrigian2021SurveySocialMH, sun2021psyqa,liu2021towards}. The potential for understanding human-human interaction patterns is limited with these single-turned or crowd-sourced datasets. ~\citet{althoff2016large} propose a multi-turn mental health counseling conversation corpus collected from a text-based crisis intervention platform, which is the best-related dataset up to now. However, the length of conversation in~\citep{althoff2016large} is shorter than ours (42 vs. 78 utterances), and the analysis mostly focuses on the counselors' utterances. In contrast, we emphasize the understanding and recognition of client reactions, which could facilitate counselors to understand the clients' feedback of their interventions as the psychological counselings proceed. 

\noindent{\textbf{Understanding Mental Health Support Conversations Using NLP.}}  Many researchers have endeavored to employ machine learning and NLP techniques to analyze mental health support conversations automatically, including modeling social factors in language that are important in the counseling setting~\citep{DBLP:journals/corr/Danescu-Niculescu-MizilSJLP13, DBLP:conf/emnlp/PeiJ20, DBLP:conf/www/0004LMAA21, DBLP:conf/naacl/HovyY21}, behavioral codes~\citep{tanana2015recursive, perez2017predicting, park2019conversation, cao2019observing}, predicting session- or utterance-level quality~\citep{gibson2016deep, goldberg2020ModelingAlliance, Wu2021RealTimeEmpathy}, and detecting mental health problems~\citep{Asad2019DepressionDetection, Xu2020InferringMH}. However, these studies again mostly focus on studying consultation skills. There are methods~\citep{tanana2015recursive,perez2017predicting} that try to classify clients' responses but only limit to a particular mental health support genre called motivational interviewing, which has an existing coding scheme with three classes for clients. Our annotation scheme is not genre specific and has more fine-grained analysis, and is more related to research in psychotherapy. 

\noindent{\textbf{Analysis of Conversation Outcome in Psychotherapy Research.}} Different from NLP research where most studies focus on the counselor side, in psychotherapy research, the interactions between counselors and clients are widely investigated \citep{ribeiro2013TCCS, norcross2010therapeutic, falkenstrom2014working}. The working alliance between the counselor and clients is a crucial researched element~\footnote{\textbf{Working Alliance}: "\textit{the alliance represents interactive, collaborative elements of the relationship (i.e., therapist and client abilities to engage in the tasks of therapy and to agree on the targets of therapy) in the context of an affective bond or positive
attachment}"~\citep{constantino2002working}.} ~\citep{norcross2010therapeutic, falkenstrom2014working}. This is because the formation of working alliance is arguably the most reliable predictor of counseling conversation outcomes~\citep{ribeiro2013TCCS}, yet it is difficult for counselors to gauge accurately during counselings. The scores of alliance rated after each counseling from therapists “appear to be independent of~\ldots alliance data obtained from their patients”~\citep{horvath1994working}. Additionally, limited by the data resource and analysis tools, most alliance analyses in psychotherapy research are either in small sample size~\citep{ribeiro2013TCCS} with only a few sessions or in session level~\citep{hatcher1999therapists}. We instead conduct a moment-by-moment analysis on a large-scale dataset and pursue an automatic solution.

\section{Annotation Framework}
\label{annotation_framework}

\begin{figure}
    \centering
    \scalebox{0.45}{
    \includegraphics{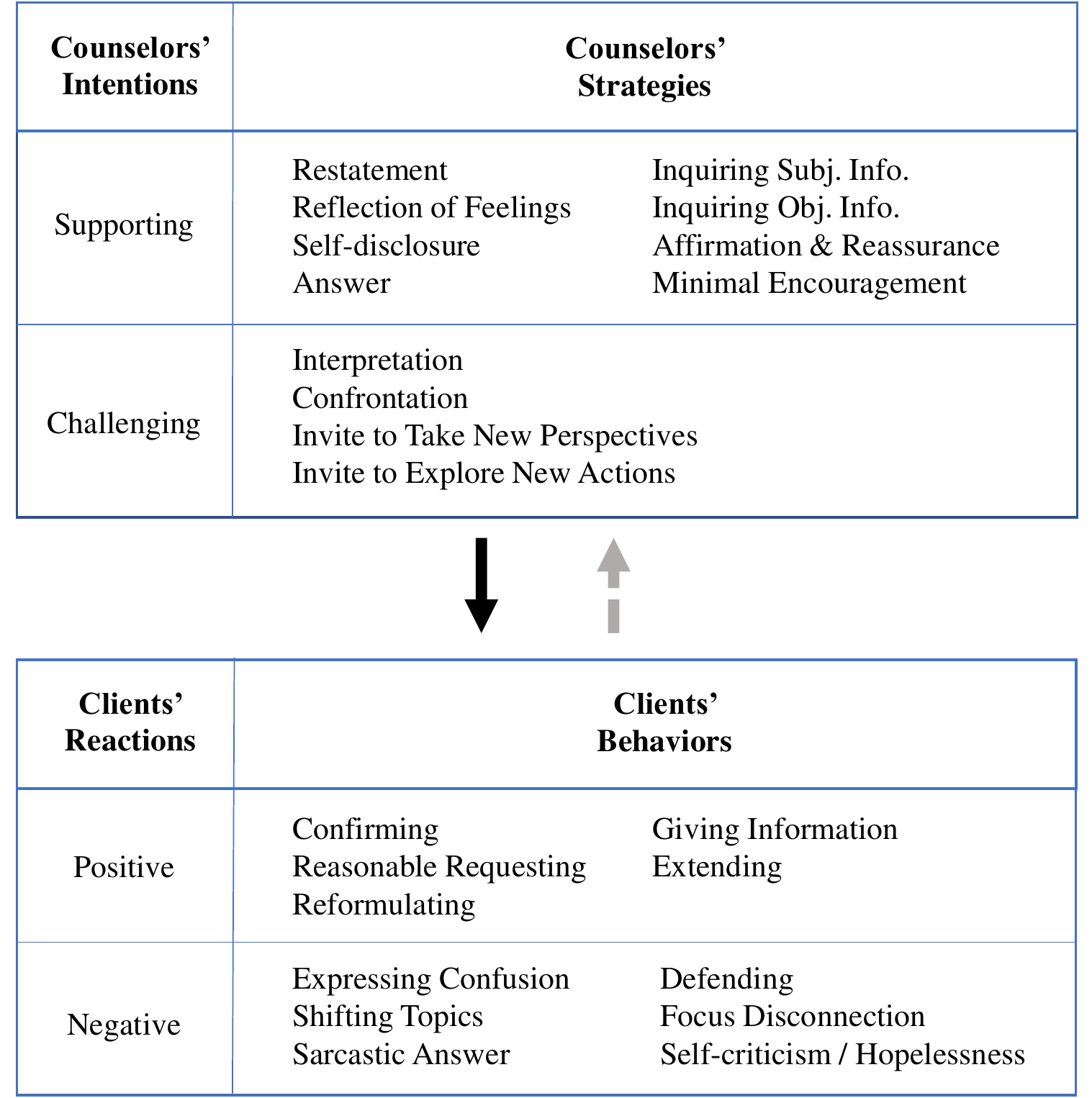}}
    \caption{Overview of our proposed framework. It contains the intentions and conversational strategies of counselors, as well as the reactions and behaviors of clients. The \textbf{black} arrow indicates the influence of counselors' intervention on clients' reactions and behaviors. The \textcolor{gray}{dashed gray} arrow indicates that clients' feedback is regarded as signals for counselors to adjust intentions and strategies in conversations.
    }
    \label{fig:framework}
\end{figure}

To understand interaction patterns between counselors and clients in text-based counseling conversations, we develop a novel framework to categorize the reactions and behaviors of clients as well as the intentions and conversational strategies of counselors (Figure~\ref{fig:framework}). In collaboration with experts in counseling psychology, we adapt and synthesize the existing face-to-face counseling-focused taxonomies, including Client Behavior System~\citep{Hill1992CBS}, Therapeutic Collaboration Coding Scheme~\citep{ribeiro2013TCCS}, Helping Skills~\citep{hill2009helping}, and Client Resistance Coding Scheme~\citep{CHAMBERLAIN1984144resistance}, to the online text-only counseling conversation settings. We have three developers~\footnote{One is a Ph.D. in psychology and a State-Certificated Class 3 Psycho-counselor with 3 years of experience; another is a State-Certificated Class 2 Psycho-counselor with more than 10 years of experience; and the last one is a doctoral student majoring in computer science and the first author of this paper.}  to carefully build the framework, following the consensual qualitative research method~\citep{hill1997CQR, ribeiro2013TCCS, park2019client-classifying}. The details of the framework development process are shown in Appendix~\ref{framework_development}. We also compare our framework with existing annotation frameworks in Appendix~\ref{comparison}.

\subsection{Counselor Intentions and Conversational Strategies}

\textbf{Counselor Intentions.}
Our taxonomy consists of two key counselor intentions, \textit{Supporting} and \textit{Challenging}, providing an outlook of how counselors orient the conversation flow ~\citep{ribeiro2013TCCS, 2020Balancing}. 

In a counseling conversation, the counselor must focus on engaging with the client’s concerns and providing an empathetic understanding~\cite {rogers1957necessary, hill2000client}. However, overemphasizing the supportive strategies might keep the client from progressing~\cite {2020Balancing, ribeiro2013TCCS}. To direct the conversation towards a positive outcome that benefits clients, the counselor should challenge and prompt the client to make some changes~\cite {mishara2007helper, 2020Balancing}. By analyzing the collected counseling conversations, we do find it common for counselors to employ supportive and challenging strategies alternatively in practice.

\noindent{\textbf{Conversational Strategies.}}
Our taxonomy contains eight \textit{Supporting} and four \textit{Challenging} fine-grained conversational strategies. We present detailed definitions and examples in Appendix~\ref{strategy_definition}.

Counselors utilize various conversational strategies to convey their intentions~\citep{hill2009helping}. To provide support, the counselors reflect on the contents or feelings the client has shared to make the client feel heard and understood (\textit{Restatement} and \textit{Reflection of Feelings}). The counselor also affirms the client's strengths or normalizes the client's negative emotions by expressing reassurance (\textit{Affirmation and Reassurance}). On the other hand, to prompt the client to make progress, the counselor might point out the client's unreasonable beliefs (\textit{Confrontation}) or encourage him or her to brainstorm solutions (\textit{Invite to Explore New Actions}). 

Notably, our annotation framework captures functional details of conversational strategies~\cite {ribeiro2013TCCS}. For example, although both \textit{Interpretation} and \textit{Invite to Take New Perspectives} encourage clients to view life from different angles, the way in which the insights are provided differs. \textit{Interpretation} strategy directly provides a new meaning, reason, or explanation to the client's behavior, thought, or emotion from the perspective beyond the client's statement or cognition. For example, "Comparing yourself to others makes you feel unsatisfied with yourself. But everyone's growth has its timeline". While \textit{Invite to Take New Perspectives} strategy usually guides the client to think from a new perspective by asking questions. For example, "If your closest friend heard your appeal, what do you think he would say to you?"

\subsection{Client Reactions and Behaviors}

\noindent{\textbf{Client Reactions.}}
The counselors' interventions elicit the clients' reactions, which is an important criterion for judging the effectiveness of counselors' previous interventions. The clients' reactions towards the counselors' interventions can be categorized as \textit{Positive} or \textit{Negative} as feedback of whether they understand counselors' purposes of using specific intentions and strategies~\citep{2001TZPD, hill2009helping, ribeiro2013TCCS}. For example, when the counselor utilizes \textit{Affirmation and Reassurance} strategy to show empathy to the client by saying, " You have a great insight into yourself!", the client may experience being understood and respond with confirmation by saying, "Thank you for your accomplishment!"; or the client may find the mere consolation is useless in resolving the dilemma of the moment and then express dissatisfaction with the counselor's intervention by saying "You always comfort me. But is there any concrete advice or suggestions?". The client's negative reactions indicate that the counselor intentions fail to achieve the intentions as expected, indicating the counselor needs to adjust strategies in the ensuing conversations~\cite {thomas1983pragmatic-failure, 2020Balancing, li2022real-time}.

\noindent{\textbf{Behaviors.}}
Our taxonomy contains five and six fine-grained behavior types for clients' \textit{Positive} and \textit{Negative} reactions, respectively. Detailed definitions are in Appendix~\ref{behavior_definition}.

Clients react to the counselor's interventions through different behaviors. For example, when the counselor provides a perspective different from a client to help the client understand a distressing experience (\textit{Interpretation}), the client may express approval (\textit{Confirming}) or start introspection (\textit{Extending}); on the contrary, the client may still insist on individual inherent views and directly express disagreement with what the counselor has said (\textit{Defending}) or show disinterest in counselor's words implicitly by changing the topic (\textit{Changing Topics}).

\section{Data Collection}
To validate the feasibility of our proposed framework in the psychological counseling conversation, we collect a large-scale counseling corpus and carefully annotate a subset of these conversations according to the framework. Our dataset will be made available for researchers who agree to follow ethical guidelines.

\subsection{Data Source}
\label{data source}

We build an online mental health support platform called Xinling to allow professional counselors to provide each client with a free text-based counseling service of about 50 minutes each time, which is a widely recognized basic time setting in psychological counseling. After each conversation, clients are asked to report their clarity on the approaches to solve existing problems by rating the conversations based on the following aspects: (1) Awareness of the changes that can be made; (2) New perspectives of looking at the problems; (3) Confidence in the ways of coping with the problems; (4) Confidence in the conversations that can lead to desirable outcomes. Clients' self-reported scores on these scales have been recognized as a consistent and major positive indicator of effective counseling~\cite {tracey1989wai, hill2009helping}. Details of the post-survey are in Table~\ref{tab:scale} in Appendix~\ref{scale_content}. We then collect counseling conversations between actual clients and experienced counselors from this counseling platform. 

In the end, we collect 2,382 conversation sessions, 479 of which receive the self-reported scales from the clients. To our knowledge, this is the largest real-world counseling conversation corpus in Mandarin. The statistics of all the collected conversations are presented in Table~\ref{tab:data_statistics}. We observe that, on average, these conversations are much longer than existing conversations collected through crowdsourcing (78.49 utterances compared to 29.8 utterances in ESConv~\citep{liu2021towards}), indicating that, in real scenarios, the professional counseling conversations contain more turns of interaction. Meanwhile, clients express longer utterances than counselors (avg. 32.48 characters compared to 24.11 characters) because clients need to give details of their problems and are encouraged to express them in the conversations, while counselors mainly act as listeners.

\begin{table}[h]
\scalebox{0.75}{
\begin{tabular}{@{}cccc@{}}
\toprule
\textbf{Category}                     & \textbf{Total} & \textbf{Counselor} & \textbf{Client} \\ \midrule
\textbf{\# Dialogues}                 & 2,382          & -                  & -               \\
\textbf{\# Dialogues with Scores}     & 479            & -
            & -               \\ 
\textbf{\# Speakers}                  & 848            & 40                 & 808             \\
\textbf{\# Utterances}                & 186,972        & 93,851             & 93,121          \\
\textbf{Avg. utterances per dialogue} & 78.49          & 39.40              & 39.09           \\
\textbf{Avg. length per utterance}    & 28.28          & 24.11              & 32.48           \\ \bottomrule
\end{tabular}}
\caption{Statistics of the overall conversations.}
\label{tab:data_statistics}
\end{table}

\subsection{Annotation Process}
\label{annotation process}
We randomly annotate a subset of sessions (520 sessions) based on the proposed framework\footnote{Before annotation, we anonymize all the client's personal information, including name, organization, etc., to protect their privacy.}. Previous research found it difficult to accurately identify counselors' conversational skills~\citep{lee2019identifying, sharma2020computational-empathy} and challenging to categorize clients' behaviors due to the linguistic diversity~\citep{lee2019identifying}. To ensure high-quality labeling, we carefully select and train 12 annotators offline. To further enhance inter-rater reliability continuously, we design a novel training-in-the-loop annotation process. The overall average inter-rater agreement on labeling counselors' and clients' utterances is 0.67 and 0.59, respectively, validating the reliability of the data. Details about the process of annotators selection and training and the training-in-the-loop policy are displayed in Appendix~\ref{annotation_process}. We use a free, open-source text annotation platform called Doccano\footnote{https://github.com/doccano/doccano} to annotate.

\subsection{Data Characteristics}
Table~\ref{tab:annotation_statistics} shows the statistics of all the annotations, including counselors' intentions and strategies and clients' reactions and behaviors.
\begin{table}[h]
\scalebox{0.7}{
\begin{tabular}{@{}clrr@{}}
\toprule
\multicolumn{1}{l}{}                              & \textbf{Categories}                                                                  & \textbf{Num} & \textbf{Mean Length} \\ \midrule
%
\multirow{14}{*}{\rotatebox{90}{\textbf{Counselors' Intentions and Strategies}}} 
                                                  & \textit{Supporting} 
                                                  & \textit{20608}      & \textit{16.80}                \\ 
                                                  & Restatement
                                                  & 4553      & 24.54                \\
                                                  & Reflection of Feelings                                                               & 729          & 20.08                \\
                                                  & Self-disclosure                                                                      & 122          & 34.5                 \\
                                                  & Inquiring Subjective Information                                                     & 5746         & 18.06               \\
                                                  & Inquiring Objective Information                                                      & 2424         & 16.20                \\
                                                  & Affirmation and Reassurance                                                          & 3279         & 17.99               \\
                                                  & Minimal Encouragement                                                                & 3485         & 2.53                 \\
                                                  & Answer                                                                               & 273          & 17.46                \\
                                                  \cmidrule(l){2-4}
                                                  & \textit{Challenging}
                                                  & \textit{5198}       & \textit{33.95}               \\ 
                                                  & Interpretation                                                                       & 2209         & 36.30                \\
                                                  & Confrontation                                                                        & 141          & 26.27                \\
                                                  & Invite to Explore New Actions  & 2495         & 33.57                \\
                                                  &  Invite to Take New  Perspectives   & 353          & 25.02                \\ 
                                                  \cmidrule(l){2-4} 
                                                  & Others                                                                               & 3593         & 17.57                \\ \cmidrule(l){2-4} 
                                                  & \textbf{Overall}                                                                    
                                                  & \textbf{29399}        &         \textbf{19.92}             \\ \midrule
\multirow{13}{*}{\rotatebox{90}{\textbf{Clients' Reactions and Behaviors}}} 
                                        & \textit{Positive}
                                        & \textit{22136}        & \textit{32.72}                                            \\
                                        & Giving Information             
                                        & 15365      & 40.91                                                              \\
                                        & Confirming                    
                                        & 3789         & 3.52                                                              \\
                                        & Reasonable Request            
                                        & 908          & 16.47                                                              \\
                                        & Extending                      
                                        & 1904        & 32.52                                                              \\
                                        & Reformulation                  
                                        & 170          & 33.12                                                             \\
                                        \cmidrule(l){2-4} 
                                        & \textit{Negative}
                                        & \textit{753}   & \textit{18.65}                                                   \\
                                        & Expressing Confusion           
                                        & 214          & 12.31                                                              \\
                                        & Defending                     
                                        & 425          & 20.72                                                              \\
                                        & Self-criticism or Hopelessness 
                                        & 51           & 17.27                                                              \\
                                        & Changing Topics                
                                        & 20           & 26.55                                                              \\
                                        & Sarcastic Answer               
                                        & 32           & 18.53                                                              \\
                                        & Focus Disconnection            
                                        & 11            & 54.45                                                             \\
                                        \cmidrule(l){2-4} 
                                        & Others                         
                                        & 3245         & 9.19                                                               \\ \cmidrule(l){2-4} 
                                        & \textbf{Overall}   & \textbf{26134}        &    \textbf{29.40}                    \\ \bottomrule

\end{tabular}}
\caption{Statistics of all the annotations, including counselors' intentions and strategies and clients' reactions and behaviors.}
\label{tab:annotation_statistics}
\end{table}

Overall, counselors use about four times more supporting than challenging strategies. The most frequently used strategy is \textit{Inquiring Subjective Information} which helps counselors gain a deeper understanding of clients' cognitive and behavioral patterns by exploring their subjective feelings, thoughts, and reasons behind them. According to challenging strategies, \textit{Confrontation} is used much less than \textit{Interpretation} and \textit{Invite to Explore New Actions}. This phenomenon is in line with the existing theory of helping skills in supportive conversations~\citep{hill2009helping} that \textit{Confrontation} should be used with caution because directly pointing out clients' incorrect beliefs or inconsistencies in conversations is likely to damage the relationship between counselors and clients.

As for clients' reactions and behaviors, clients' \textit{Positive} reactions towards counselors' interventions are significantly more than the \textit{Negative} ones, demonstrating an overall high quality of the collected counseling conversations. The most frequent behavior is \textit{Giving Information}, which corresponds to the amount of counselors' strategy \textit{Inquiring Subjective and Objective Information}, the clients provide the information that the counselors ask for. Besides, \textit{Defending} is the most common negative behavior, reflecting that counselors try to get clients to change their perspectives or behaviors during conversations. Still, clients feel hard to follow and therefore defend and insist on their original cognitive and behavioral patterns. Some more extreme behaviors, such as \textit{Self-criticism or Hopelessness}, rarely occurs, hence post difficulties for us to understand these behaviors and build good classifiers on them. 

\section{Application to Online Counseling}
\label{application}
To illustrate how the proposed framework can be used to monitor and improve the effectiveness of conversations, we conduct the following analyses:

First, we demonstrate clients' positive and negative reactions and behaviors affect the final counseling effectiveness (Section~\ref{validation_client}). We then show how clients react to counselors' intentions and strategies (Section~\ref{validation_counselor}). Based on these findings, we investigate how counselors can adjust their strategies accordingly to make entire conversations more effective (Section~\ref{insights}). Finally, we build a baseline model for automatically labeling each counseling strategy and client behavior (Section~\ref{automatic_prediction}).

\subsection{How Client Reactions Indicates Counseling Effectiveness}
\label{validation_client}
 To derive a simple conversation-level measurement, we calculate the proportion of each reaction or behavior over all the client messages in a conversation. We use the client's perceived total score on the post-conversation survey as an effectiveness indicator.

\begin{figure}[ht]
    \centering
    \scalebox{0.2}{
    \includegraphics{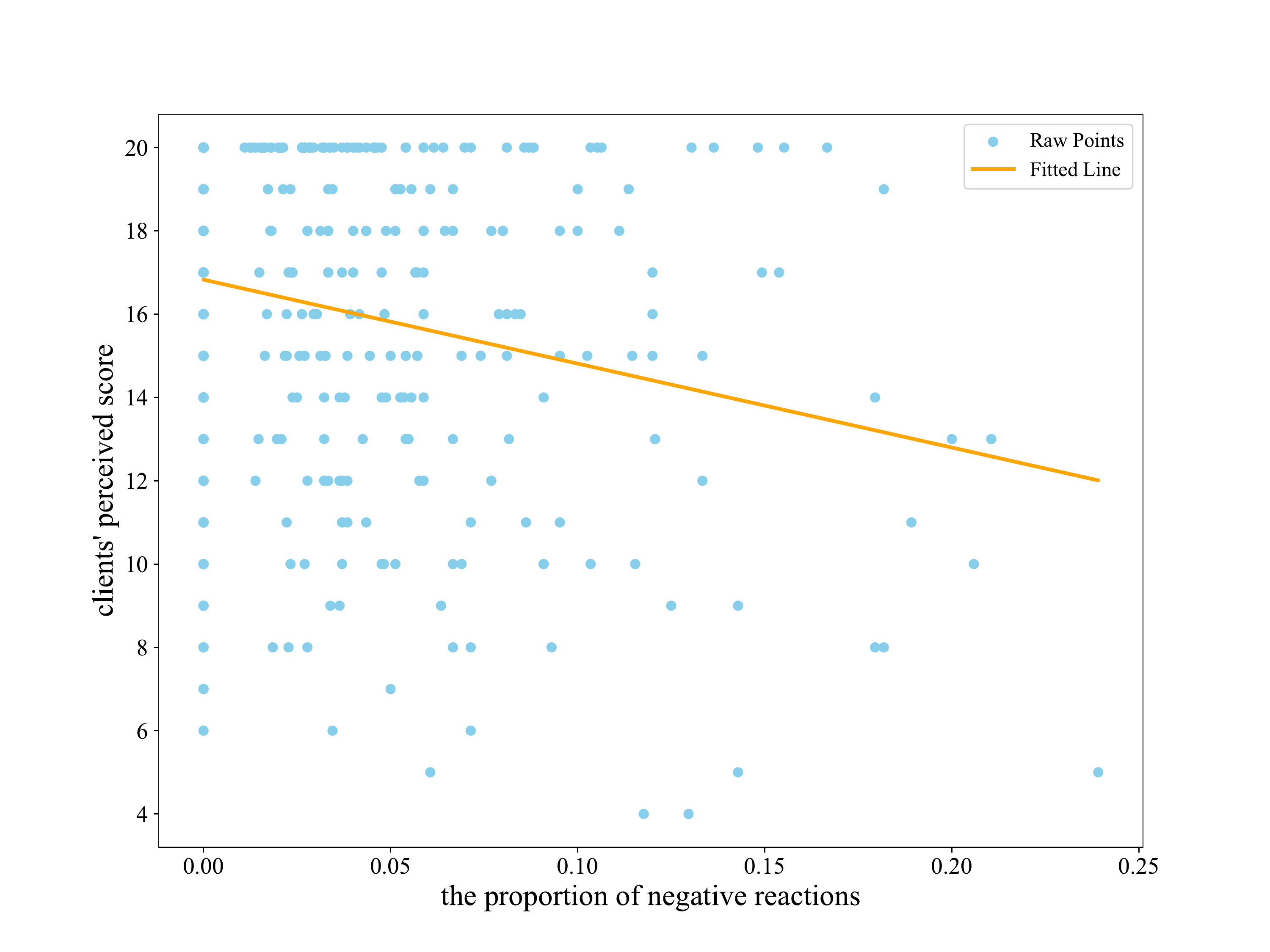}}
    \caption{The scatter plot of clients' negative reactions and their perceived conversation-level scores (the blue points) with the best-fit line (the orange line).}
    \label{fig:scatter_with_best_fit_line}
\end{figure}

\paragraph{Reactions} The relationship between the distribution of negative reaction types and client-rated conversation effectiveness is analyzed by Pearson Correlation Analysis~\citep{lee1988correlation_coefficient}. The results show that the proportion of the clients' negative reactions and the conversation effectiveness correlate negatively with  correlation coefficient $\rho = -0.2080$ and p-value $p = 1.7591e^{-5}$. Specifically, when clients have more \textit{Negative} reactions to counselors' interventions, they give a lower score of conversation effectiveness (see Figure~\ref{fig:scatter_with_best_fit_line}). The findings echo the definition of clients' \textit{Negative} reaction types, which place a negative impact on the effectiveness of counseling conversations.



\paragraph{Behaviors}
In order to find out the behaviors that influence conversation effectiveness the most, we fit a lasso model  with the proportion of the client's each behavior type as independent variables and the scores of conversation effectiveness as the dependent variable. In the end, we find that the most influential positive and negative behaviors are \textit{Extending} and \textit{Defending} (Detailed results of the importance of each behavior are in Appendix \ref{important_beavhior}), respectively. which is in line with the fact that counseling conversations are more likely to be effective when clients perceive themselves in a new way or experience changes in their behaviors, thoughts, or feelings but to be less effective when clients defend their mistaken belief~\citep{Hill1992CBS, ribeiro2013TCCS}.

To further understand the effect of negative behaviors on conversation effectiveness, the average score of the conversations with at least one negative behavior is calculated, which is 15.79, a drop of about 2\% from the overall average score (Table~\ref{tab:negative_occurrence_avg_score}). 
The results  again indicate that clients' negative behaviors harm conversation effectiveness. Notably, \textit{Defending} happens in most of the sessions that have negative behaviors. The overall low scores with defending behavior indicate that the conversation effectiveness is damaged when the clients start to defend and insist on their original beliefs. Although other negative behaviors such as \textit{Changing Topics} have lower overall scores, they happen in fewer sessions and are less influential in our dataset. Once we have enough data for these categories, we expect their importance to become more apparent.

\begin{table}[ht]
\normalsize
\scalebox{0.73}{
\begin{tabular}{@{}lcr@{}}
\toprule
\textbf{Categories}                                   & \textbf{Avg. Score} & \textbf{\# Sessions} \\ \midrule
Changing Topics                                       & 14.57 & 14   \\
Sarcastic Answer                                      & 14.40 & 10   \\
Focus Disconnection                                   & 13.25 & 4    \\
Defending                                             & 15.46 & 175  \\
Self-criticism or Hopelessness                        & 14.04 & 24   \\
Expressing Confusion                                  & 16.05 & 127  \\ \midrule
All Conversations                                     & 16.14 &  419  \\
Conversations with Negative Behaviors                 & 15.79 &  239  \\ \bottomrule
\end{tabular}}
\caption{The effect of the occurrence of each negative behavior on the conversation effectiveness.}
\label{tab:negative_occurrence_avg_score}
\end{table}


\subsection{Similar Counseling Strategies Leads to Similar Client Reactions}
\label{validation_counselor}

The clients react and behave differently towards counselors' different strategies. We find that counselors' strategies with the same intention lead to similar clients' behaviors. Specifically, strategies belonging to \textit{Challenging} result in a larger proportion of clients' follow-up \textit{Negative} behaviors than those belonging to \textit{Supporting} (4.77\% vs. 2.87\%). The findings verify the rationality of categorizing the counselors' strategies into \textit{Supporting} and \textit{Challenging}. The detailed analysis is shown in Appendix~\ref{clients_behavior_distribution}.

\begin{figure}[ht]
    \centering
    \scalebox{0.25}{\includegraphics{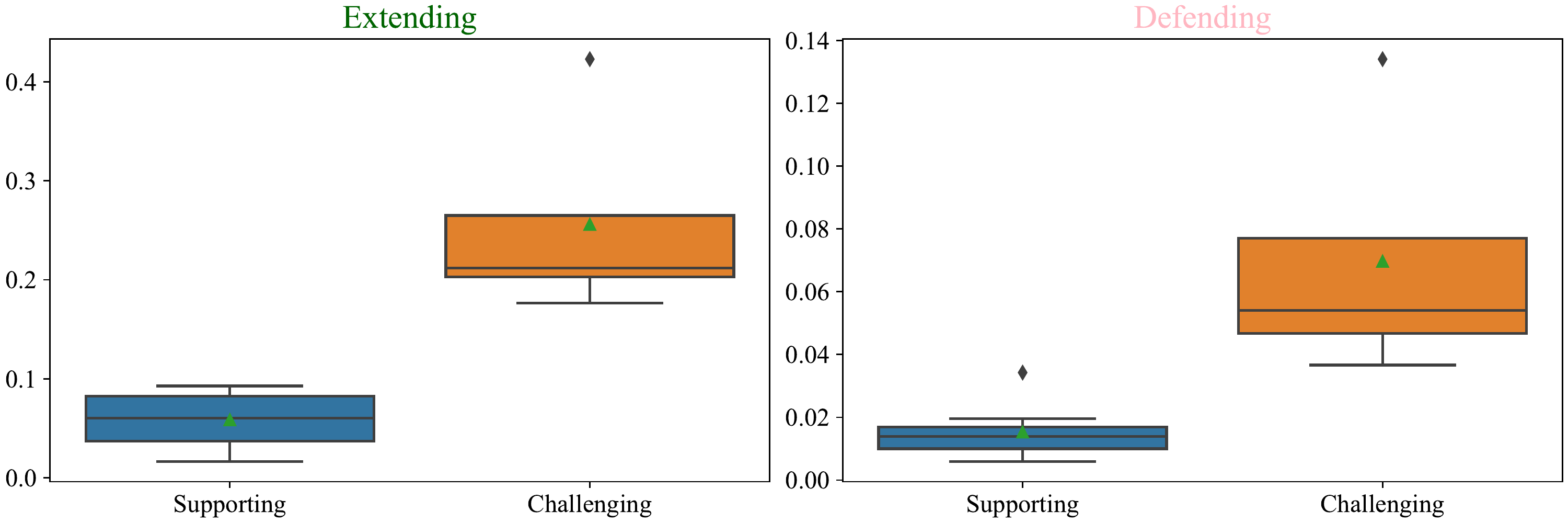}}
    \caption{The distribution of clients' \textit{Extending} and \textit{Defending} behaviors after the counselors' \textit{Supporting} and \textit{Challenging} strategies.}
    \label{fig:intentions2behaviors_boxplot}
\end{figure}

We then explore the influence of the counselors' strategies of \textit{Supporting} and \textit{Challenging} on clients' \textit{Extending} and \textit{Defending} behaviors as these are the most important ones according to the above analysis. As shown in Figure~\ref{fig:intentions2behaviors_boxplot}, compared with the \textit{Supporting}, the \textit{Challenging} brings a higher proportion of the clients' \textit{Extending} behaviors. Meanwhile, \textit{Challenging} makes the clients \textit{Defending} as well. Therefore, to improve the conversation effectiveness, the appropriate utilization of the counselors' \textit{Challenging} strategies is important, and we will analyze it in the following section.


\subsection{Appropriate Strategy Utilization} 

\label{insights}

To explore how counselors utilize \textit{Challenging} appropriately to make clients behave as \textit{Extending}, rather than \textit{Defending}. We focus on two factors that influence the effectiveness of strategies: conversation stages and interaction patterns in the conversation history between counselors and clients\citep{althoff2016large}.

\begin{figure}[ht]
    \centering
    \scalebox{0.25}{\includegraphics{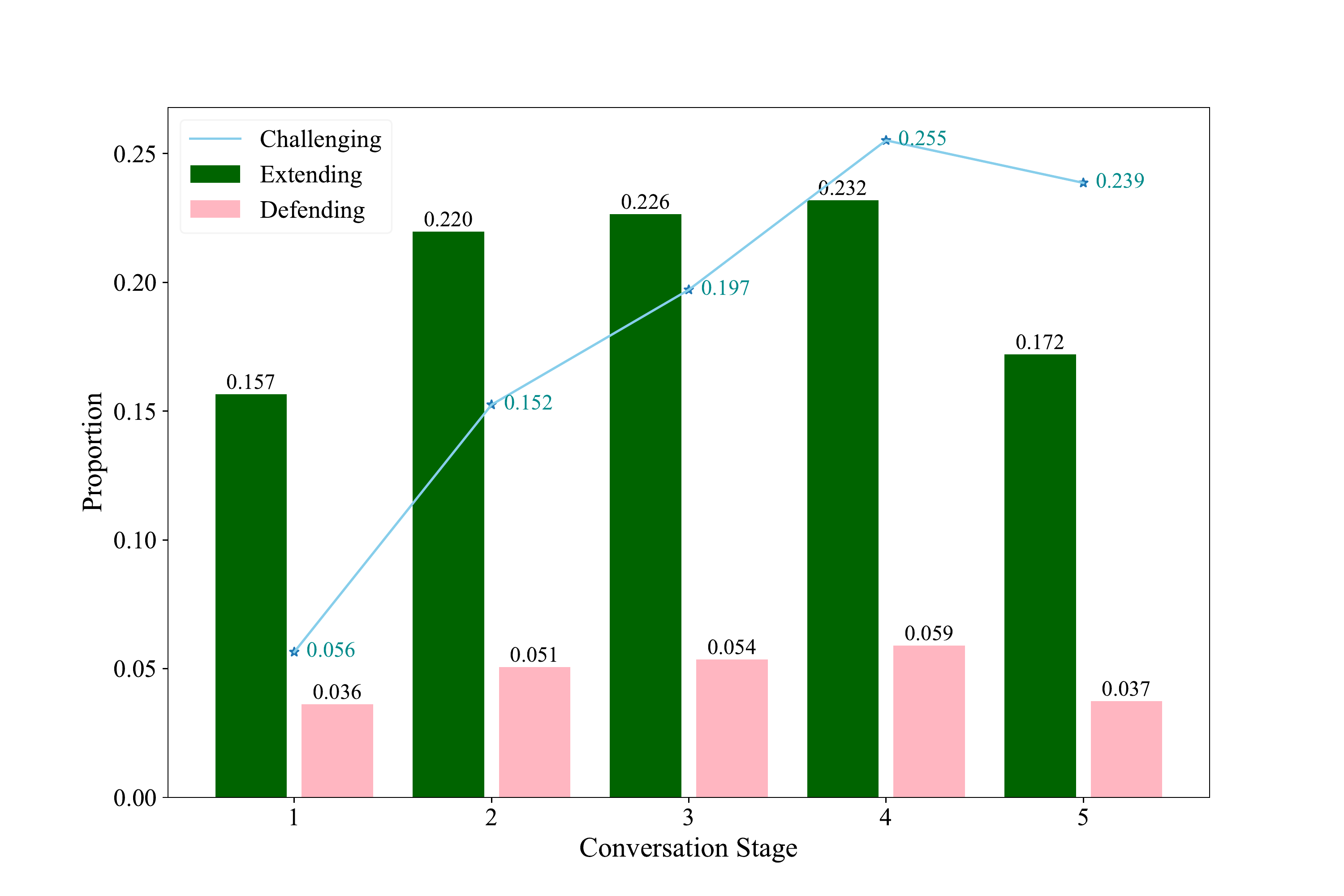}}
    \caption{The distribution of clients' \textit{Extending} and \textit{Defending} reactions after counselors' \textit{Challenging} strategy at different stages. The sky-blue line presents the proportion of \textit{Challenging} strategy utilized in each stage.}
    \label{fig:challenging_intentions_by_coarse_strategies2stage_with_behavior}
\end{figure}

\noindent{\textbf{Conversation Stages.}} Each conversation is divided uniformly into five stages,  and the distribution of clients' certain behaviors after counselors' \textit{Challenging} at each stage is computed. Due to the high proportion of content in the first and last stages (18.70\% and 33.86\%) being irrelevant to counseling topics (labeled as \textit{Others}), only the content in the middle three stages are analyzed. As shown in Figure~\ref{fig:challenging_intentions_by_coarse_strategies2stage_with_behavior}, the counselors utilize more and more \textit{challenging} as the conversations progress. Meanwhile, both \textit{Extending} and \textit{Defending} increase when clients face counselors’ \textit{Challenging}. Since \textit{Extending} is overall more common than \textit{Defending}, this phenomenon suggests that counselors adopt \textit{Challenging} step by step within a counseling session. We will leave the cross-section analysis in future work. 



\noindent{\textbf{Counselor-Client Preceding Interaction Patterns.}} The counselor-client preceding interaction is defined as the pair of the counselors' \textit{Supporting} or \textit{Challenging} and the clients' following-up \textit{Positive} or \textit{Negative} reactions. We fit a logistic regression classifier to study how these preceding interaction patterns affect the \textit{Extending} and \textit{Defending} behaviors when facing a \textit{Challenging} strategy. The overall classification accuracy is around 80\%, but we care more about the fitted coefficients, shown in Table~\ref{tab:interaction_patterns_logistic_regression}. As can be seen, if the clients reacted positively to the counselors' \textit{Challenging} before, the probability of the clients' \textit{Extending} reactions increase when the counselors intervene with \textit{Challenging} again, and vice versa. In other words, if counselors detect negative reactions from their clients, especially because of their supporting strategy, they should address those issues before launching into challenging strategies. In the event that they challenge their clients and receive positive reactions, they can continue to use the same strategy. 


\begin{table}[ht]
\centering
\resizebox{0.6\linewidth}{!}{
\begin{tabular}{@{}lr@{}}
\toprule
\textbf{Interaction Patterns}   &   \textbf{Coefficients}             \\ \midrule
Supporting - Positive   &     1.3041***                        \\
Supporting - Negative   &   -9.0643***                         \\
Challenging - Positive  &   3.7189***                         \\
Challenging - Negative  &    -7.3665***                        \\ \bottomrule
\end{tabular}}
\caption{Associations between the counselor-client interaction patterns in preceding conversations and clients' current behaviors in response to counselors' \textit{Challenging} interventions. ***$p$ \textless $0.001$. The coefficients are from a logistic regression predicting the probability that the clients behave as \textit{Extending} rather than \textit{Defending}.}
\label{tab:interaction_patterns_logistic_regression}
\end{table}

\subsection{Baseline Classifiers for Automatic Label Prediction}
\label{automatic_prediction}

To facilitate counselors guessing their clients’ states, we train classifiers to categorize counselors' intentions and strategies and identify clients' reactions and behaviors based on a pre-trained Chinese RoBERTa-large model~\citep{cui-etal-2020-revisiting}. Each task assigns a label to each sentence in a long utterance, utilizing conversation history as the context. To improve the domain adaption of pre-trained models~\citep{2020Don, sharma2020computational-empathy}, we perform the masked language modeling (MLM) task on all the collected conversations and then jointly train each classification task on the annotated data with the MLM as an auxiliary task. More experimental details are shown in Appendix~\ref{experimental_details}.

 As shown in Table~\ref{tab:classification_results}, the test set of four tasks. The model's performance in categorizing counselors' intentions and strategies is better than identifying clients' reactions and behaviors. The overall performance on identifying clients' reactions is limited by \textit{Negative} reactions (F1-value = 34.78\%). The results indicate that clients' reactions are difficult to identify, especially the negative behaviors~\cite {lee2019identifying, cao2019observing}.

The major error in predicting clients' behaviors comes from the confusing \textit{Reformulating} with \textit{Extending}. In both cases, the client is making changes, but the former changes more deeply. Besides, \textit{Defending} is hard to identify due to clients' diverse expressions of resistance. Clients may defend themselves by expressing different opinions from counselors rather than directly denying them, which is difficult for the model to recognize. More detailed classification results are in Appendix~\ref{experimental_results}. 

\begin{table}[ht]
\scalebox{0.7}{
\begin{tabular}{@{}ccccc@{}}
\toprule
\textbf{Task} & \textbf{Acc.} & \textbf{Precision} & \textbf{Recall} & \textbf{Macro-F1} \\
\midrule
 Intentions   & $0.9025_{0.0030}$  & $0.8821_{0.0046}$  & $0.8446_{0.0045}$  & $0.8612_{0.0040}$   \\ 
 Strategies  & $0.8103_{0.0035}$  &  $0.7317_{0.0236}$ & $0.6533_{0.0082}$ & $0.6791_{0.0074}$\\
\midrule
Reactions   &  $0.9490_{0.0016}$ & $0.7762_{0.0163}$ & $0.6977_{0.0167}$ & $0.7214_{0.0138}$ \\
Behaviors    & $0.8597_{0.0018}$ & $0.5815_{0.0273}$ & $0.5190_{0.0140}$ & $0.5354_{0.0155}$ \\

\bottomrule
\end{tabular}}
\caption{The overall results of the test set of four tasks: categorizing counselors' intentions and strategies, and clients' reactions and behaviors (Due to the scarce number of \textit{Changing Topics}, \textit{Sarcastic Answer} and \textit{Focus Disconnection}, we filter out these samples when building classifiers). We report averages across five random seeds, with standard deviations as subscripts.}
\label{tab:classification_results}
\end{table}

\section{Conclusion}

We develop a theoretical-grounded annotation framework to understand counselors' strategies and clients' behaviors in counseling conversations. Based on a large-scale and high-quality text-based counseling dataset we collected over the past two years, we validate the plausibility of our framework. With the labeled data, we also find that clients' positive reactions boost their ratings of counseling effectiveness, but negative reactions undermine them. Meanwhile, clients are more likely to \textit{extend} after counselor \textit{challenge} their beliefs. Moreover, our automatic annotation models indicate that clients' reactions and behaviors are more difficult to predict than counselors' intentions and strategies.  Due to the complexity of the data and the lack of labeled data for rare cases, our analysis is relatively shallow. We analyze the weakness of our work in section \ref{limitation} and will dig deeper into each interaction pattern once we have more data. 

\section{Limitations}
\label{limitation}
As this is the first large-scale analysis of client reactions in online mental health counseling, there is huge room for future improvement. Here we only list a few problems that we would like to address in the short future. First, although our annotation framework is comprehensive, the data labeled is quite imbalanced. In some rare classes, there are fewer than 50 instances, making it difficult to conduct an in-depth analysis, let alone train an accurate classifier. Therefore, our analysis mostly focuses on the \textit{Extending} and \textit{Defending} behaviors. We will label more data so that rare cases can be better understood and classified more accurately. The accuracy of a classifier is important for real-life applications because it has the potential to mislead counselors. Second, we only have one short post-survey, which limits our coarse-scale analysis. We are adding more and richer post-surveys. Third, while we hope that the lessons learned can be applied to everyday conversations, our analysis has only been limited to psycho-counseling. The lessons learned will be tested against a wider range of use cases. It is important, however, not to overgeneralize our findings as this may harm the naturalness of our daily conversations. After all, the psycho-counseling process is a very special type of conversation.

\section*{Acknowledgements}
We are grateful to all counselors and clients for agreeing to use their counseling conversations for scientific research, and all annotators for their hard work. We appreciate the engineers who operate and maintain the counseling and annotation platform. Besides, we would like to express our gratitude to Professor Zhou Yu and other teachers, colleagues and anonymous reviewers who provided insightful feedback and suggestions for this project.

\section*{Ethics Statement}
The study is granted ethics approval from the Institutional Ethics Committee (20211013LZZ001). All the clients and counselors signed a consent form when using our counseling platform, which informed them that the counseling conversations collected on the platform would be used for scientific research purposes, and might be used for scientific research by third parties. During the annotation process, we spared no efforts to manually de-identify and anonymize the data to protect clients’ and counselors’ privacy. The annotators also signed data confidentiality agreements and acquired ethical guidelines before they got access to the conversation data. Meanwhile, they were paid a reasonable wage for annotation. For the rules of releasing data, the third-party researchers who require access to the raw conversation data must provide us their valid ID, proof of work, the reason they request data (e.g., the research questions), etc. They are required to be affiliated with an non-profit academic or research institution. This includes obtaining the approval of an Institutional Review Board (IRB), having principal investigators working full-time as well as the written approval of institution’s office of Research or equivalent office. Additionally, they must sign the Data Non-disclosure Agreement and make promise that they would not share the data with anyone.

\bibliography{custom}
\bibliographystyle{acl_natbib}

\appendix
\label{appendix}

\section{Annotation Framework}

\subsection{Framework Development Process}
\label{framework_development}
We have three main taxonomy developers (two are experienced with clinical and emotional support, and one is the first author) to develop the framework, following the consensual qualitative research method~\citep{hill1997CQR, ribeiro2013TCCS, park2019client-classifying}. Here, we describe the detailed developing process for the counselor's taxonomy as an example.

Firstly, based on existing taxonomies~\citep{ribeiro2013TCCS, hill2009helping}, we filter those categories that are not appropriate for the text-only conversation settings (e.g., silence, head nods) and create the first version of taxonomy and annotation guideline. Secondly, we randomly select 6-10 conversations and ask all the taxonomy developers to annotate them independently. After the annotation, the developers discuss the differences and confusions among their annotations until reaching a consensus. During this process, they may add, merge or delete certain categories and refine the annotation guideline. We repeat the above step two for five times to obtain the final version of the taxonomy and guideline, including detailed definitions and examples. The Fleiss kappa values~\citep{fleiss1971measuring} among the three taxonomy developers in the five iterations are as follows: 0.6255, 0.6721, 0.6819, 0.7085, and 0.7233. The monotonically increasing agreement proves that the iterative process effectively resolves differences among developers. And the substantial agreement ensures the reliability of our taxonomy. During the whole process, we annotate 30 conversations.

\subsection{Comparison with Existing Frameworks}
\label{comparison}
We compare our proposed framework with existing annotation frameworks for analyzing dialogue acts of participants in the counseling conversations (see Table~\ref{tab:comparison}). Much research has mostly focused on studying counselors' strategies, such as CCU~\citep{lee2019identifying} and ESC~\citep{liu2021towards}. Specifically, the ESC framework proposes 7 counselors' support strategies based on three counseling stages. Different from ESC, our framework contains a more comprehensive and finer-grained classification (12 strategies) of counselors' skills based on their intentions. There are methods that attempt to classify clients' responses~\citep{park2019client-classifying, tanana2015recursive, perez2017predicting}. ~\citet{park2019client-classifying} build a novel Categorization scheme of Client Utterances (CCU) with 5 categories. Such a scheme does not contain clients' immediate feedback on counselors' interventions, especially the negative one, limiting its role in helping counselors adjust their strategies and evaluating counseling effectiveness. In ~\citep{tanana2015recursive, perez2017predicting}, researchers conduct categorization on both counselor and client sides based on MISC framework, but they are only limited to a particular mental health support genre called motivational interviewing. Our annotation framework is not genre specific and has more fine-grained analysis.

\begin{table}[ht]
    \centering
    \scalebox{0.7}{
    \begin{tabular}{c|c|c|c}
    \toprule
    \multirow{2}{*}{\textbf{Framework}}   &  \multicolumn{2}{c|}{\textbf{Categorization}} & \multirow{2}{*}{\textbf{\makecell[c]{Not \\ Genre-Specific}}}      \\ \cline{2-3}
        & \textbf{Counselor} & \textbf{Client} & \\ 
    \midrule
    \makecell[c]{CCU \\ ~\citep{park2019client-classifying}}     &   & $\checkmark$ & $\checkmark$ \\ \hline
    \makecell[c]{TCA \\ ~\citep{lee2019identifying}} & $\checkmark$ & & $\checkmark$ \\ \hline
    \makecell[c]{ESC \\ ~\citep{liu2021towards}} & $\checkmark$ &  & $\checkmark$ \\ \hline
    \makecell[c]{MISC \\ ~\citep{tanana2015recursive} \\ ~\citep{perez2017predicting}} & $\checkmark$ & $\checkmark$ & \\ \hline
    \multirow{2}{*}{Our Framework} & \multirow{2}{*}{$\checkmark$} & \multirow{2}{*}{$\checkmark$} & \multirow{2}{*}{$\checkmark$} \\ 
    & & & \\ \bottomrule
    \end{tabular}}
    \caption{A comparison of our proposed framework with other existing annotation frameworks.}
    \label{tab:comparison}
\end{table}

\subsection{Definitions of Strategies}
\label{strategy_definition}

\noindent\textbf{Restatement.} The counselor reflects the content and meaning expressed in the client's statements in order to obtain explicit or implicit feedback from the client.

\noindent\textbf{Reflection of Feelings.} The counselor uses tentative or affirmative sentence patterns to explicitly reflect the client's mood, feelings, or emotional states.

\noindent\textbf{Self-disclosure.} The counselor discloses personal information to the client, including but not limited to the counselor's own similar experiences, feelings, behaviors, thoughts, etc.

\noindent\textbf{Inquiring Subjective Information.} The counselor explores the client's subjective experience, including thoughts, feelings, states, the purpose of doing something, etc.

\noindent\textbf{Inquiring Objective Information.} The counselor asks the client to concretize the imprecise factual information, including details of events, basic information about the client, etc. 

\noindent\textbf{Affirmation and Reassurance.} The counselor affirms the client's strengths, motivations, and abilities, and normalizes the client's emotions and motivations, and provides comfort, encouragement, and reinforcement.

\noindent\textbf{Minimal Encouragement.} The counselor offers minimal encouragement to the client in an affirmative or questioning manner, encouraging the counselor to continue talking and facilitating the conversation.

\noindent\textbf{Answer.} The counselor answers the questions that the client asks about the conversation topics.

\noindent\textbf{Interpretation.} The counselor gives a new meaning, reason, and explanation to the behaviors, thoughts, or emotions of the client from a perspective beyond the client's statements or cognition, and tries to make the client look at problems from a new perspective.

\noindent\textbf{Confrontation.} The counselor points out the client's maladaptive beliefs and ideas, inconsistencies in the statements, or contradictions that the client is unaware of or unwilling to change.

\noindent\textbf{Invite to Take New Perspectives.} The counselor invites the client to use an alternative perspective to understand the given experience.

\noindent\textbf{Invite to Explore New Actions.} The counselor asks questions to guide the client to think and explore how to take new actions or invites the client to act in different ways during or after the conversation.

\subsection{Definitions of Behaviors}
\label{behavior_definition}

\noindent\textbf{Confirming.} The client understands or agrees with what the counselor has said.

\noindent\textbf{Giving Information.} The client provides information according to the specific request of the counselor.

\noindent\textbf{Reasonable Request.} The client attempts to obtain clarification, understanding, information, or advice and opinions from the counselor.

\noindent\textbf{Extending.} The client not only agrees to the counselor's intervention, but also provides a more in-depth description of the topic being discussed, including the client's analysis, discussion, or reflection on his or her original cognition, thoughts, or behaviors.

\noindent\textbf{Reformulating.} The client responds to and introspects the counselor's intervention while proposing his or her own perspectives, directions of thinking, or new behavioral patterns on current issues.

\noindent\textbf{Expressing Confusion.} The client expresses confusion or incomprehension of the counselor's intervention or directly states that he or she has no way to answer or respond to the questions or interventions raised by the counselor.

\noindent\textbf{Defending.} The client is stubborn about an experience, glorifies or makes unreasonable justifications for his or her own views, thoughts, feelings, or behaviors, and insists on seeing the experience from the original perspective.

\noindent\textbf{Self-criticism or Hopelessness.} The client falls into self-criticism or self-reproach, is engulfed in a state of desperation and expresses his or her inability to make changes.

\noindent\textbf{Shifting Topics.} Faced with the intervention of the counselor, the client's reply does not postpone the previous issue, but shifts to other issues.

\noindent\textbf{Focus Disconnection.} The client disengages from what the counselor is discussing, focuses on stating issues of interest, and does not respond to the counselor's intervention.

\noindent\textbf{Sarcastic Answer.} The client expresses dissatisfaction with the counselor, and questions or ridicules the counselor's intervention.


\section{Annotation Process}
\label{annotation_process}

\subsection{Post-survey Scales}
\label{scale_content} 

To facilitate readers understand clients' self-report results of counseling conversations in our data, we present the questions of the assessment in Table~\ref{tab:scale}. For each question, clients are required to choose only one from the following five options: seldom, sometimes, often, very often, and always, representing 1 to 5 points, respectively.


\begin{table}[ht]
\centering
\scalebox{0.75}{
    \centering
    \begin{tabular}{c|c}
    \toprule
    \textbf{No.}   &  \textbf{Questions} \\ \midrule
    1 & \makecell[l]{As a result of this session, I am clearer as to how I \\ might be able to change.} \\ \hline
    2 & \makecell[l]{What I am doing in the counseling gives me new \\ ways of looking at my problem.} \\ \hline
    3 & \makecell[l]{I feel that the things I do in the counseling will help \\ me to accomplish the changes that I want.} \\ \hline
    4 & \makecell[l]{I believe the way we are working with my problem \\ is correct.} \\ \bottomrule
    \end{tabular}}
    \caption{Questions of assessment after the counseling}
    \label{tab:scale}
\end{table}

\subsection{Annotators Selection and Training}
\label{annotators}

\noindent{\textbf{Annotators Selection and Training.}} We select 30 candidates out of more than 100 applicants who are at least undergraduate in psychology or with practical experience in counseling to attend an offline interview. During the interview, all the candidates are asked to learn the annotation guideline and then take three exams. Each exam consists of 50\textasciitilde 60 conversation snippets. For each snippet, candidates are required to annotate the last utterance. After each exam, we provide the candidates the annotations to which they assigned incorrect labels in the exam and the corresponding correct labels to help them better understand the guideline. After the interview, the top 12 candidates with the highest average accuracy on the three exams become the final annotators. The highest and lowest accuracies are 72.07\% and 64.01\%, respectively (refer to Table~\ref{tab:exam} for more details). We then conduct two-day offline training for these qualified annotators. During training, all the annotators first annotate three conversations simultaneously (305 utterances), which have a ground truth labeled by our psychological experts. Then, the annotators analyze the utterances mislabeled in group meetings.

\begin{table}[ht]
\centering
\scalebox{0.7}{
\begin{tabular}{@{}ccccc@{}}
\toprule
\textbf{\makecell{Annotator \\ ID}} & \textbf{Exam1} & \textbf{Exam2} & \textbf{Exam3} & \textbf{Avg.}                \\ \midrule
1                      & 0.6914        & 0.7582        & 0.7126        & $0.7208_{0.0279}$                \\
2                      & 0.6420        & 0.7692        & 0.7356        & $0.7156_{0.0538}$                 \\
3                      & 0.6790        & 0.7692        & 0.6552        & $0.7011_{0.0491}$                 \\
4                      & 0.5679        & 0.7802        & 0.7356        & $0.6946_{0.0914}$                 \\
5                      & 0.6296        & 0.7033        & 0.7356        & $0.6895_{0.0444}$                 \\
6                      & 0.6173        & 0.7253        & 0.7241        & $0.6889_{0.0506}$                 \\
7                      & 0.6296        & 0.6923        & 0.7356        & $0.6859_{0.0435}$                 \\
8                      & 0.6790        & 0.7143        & 0.6552        & $0.6828_{0.0243}$                 \\
9                      & 0.7161        & 0.6593        & 0.6667        & $0.6807_{0.0252}$                 \\
10                     & 0.5679        & 0.7142        & 0.7356        & $0.6726_{0.0745}$                 \\
11                     & 0.5679        & 0.7143        & 0.6782        & $0.6535_{0.0623}$                 \\
12                     & 0.6296        & 0.6813        & 0.6092        & $0.6401_{0.0304}$                 \\ \bottomrule
\end{tabular}}
\caption{The results of each and average accuracy of the selected top twelve annotators in the three exams, with standard deviations as subscripts in the last column.}
\label{tab:exam}
\end{table}

\begin{figure}[ht]
\scalebox{0.25}{
    \includegraphics{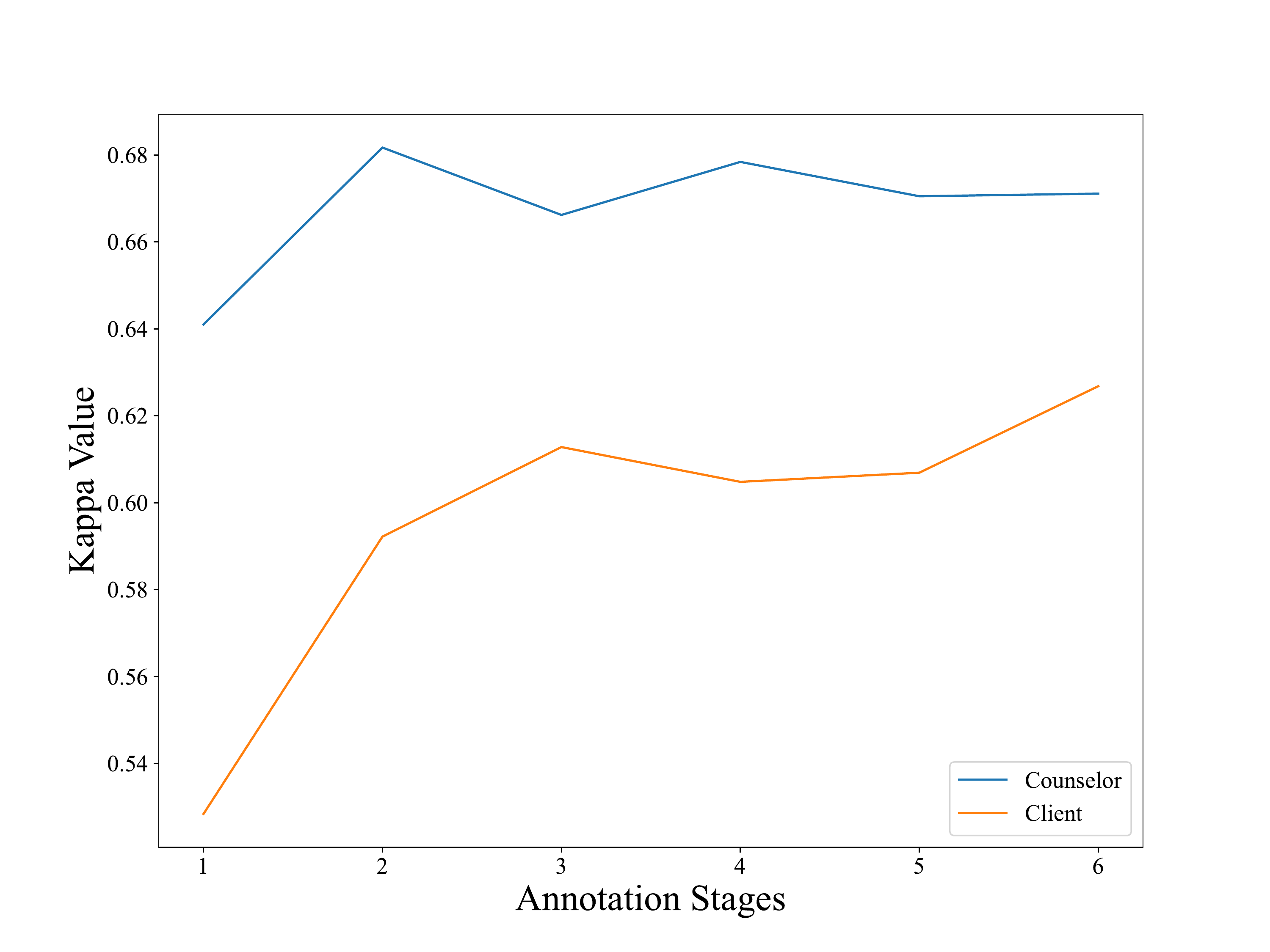}}
    \caption{Agreement among annotators on labeling counselors and clients' utterances in each annotation stage.}
    \label{fig:agreement_in_each_stage}
\end{figure}

\noindent{\textbf{Training in the Loop.}}
To further improve the inter-annotator agreement and annotation accuracy, we design the annotation process into six annotation and training stages. In the annotation stage, annotators are asked to record the utterances difficult to label (confusion samples). In the training stage, the psychological experts train each annotator after reviewing the confusing samples (618 samples) in a questions-and-answers document. As shown in Figure~\ref{fig:agreement_in_each_stage}, the average agreement of the latter stages is higher than the former stages, indicating that the training-in-the-loop policy is effective. 


\subsection{Data Quality Control}
We randomly assign each conversation to three or more annotators and ask them to annotate based on counselors' fine-grained conversational skills and clients' behavior types at the sentence level. Once obtaining the annotated data, we calculate the Fleiss' kappa~\citep{fleiss1971measuring} among multiple annotators in each conversation, which measures the proportion of agreement over and above the agreement expected by chance instead of measuring the overall proportion of agreement. For Fleiss' kappa, 0.61$\sim$0.80 is indicated as substantial agreement. Considering the task demand that we have 12 annotators who annotated 13 and 12 categories of counselors' strategies and clients' behaviors and reality of time, we take the substantial level of agreement. Finally, the average inter-rater agreement on labeling counselors' and clients' utterances is 0.67 and 0.59 respectively, validating the reliability of the data. And we find that human annotators struggle with some specific categories, such as \textit{Interpretation} versus \textit{Invite to Take New Actions} in counselors' strategies, \textit{Extending} versus \textit{Reformulation} in clients' behaviors, etc. We then utilize a majority vote method to obtain the final labels. For those samples that haven't been labeled by the above method process, we randomly assign them to three or more annotators until we get a majority vote. Overall, we find that compared to annotating counselors' conversational skills, identifying clients' reactions and behaviors is more difficult because they do not act within theoretical frameworks~\cite {lee2019identifying}.

\section{Automatic Prediction}
\label{prediction}

\subsection{Experimental Details} 
\label{experimental_details}
\paragraph{Data Preparation} Tasks for both speakers share the same data preparation process. We randomly split the annotated data into a training set (70$\%$), validation set (15$\%$), and test set (15$\%$). Note in the split, all utterances in a conversation are assigned to the same set.

\paragraph{Experimental Settings} All the models are implemented with PyTorch deep learning package~\citep{2019PyTorch}. To make the pretrained model aware of the speaker's information in conversation, we adopt a simple, special tokens strategy by prefixing a special token [SP] or [SK] to each utterance from counselors or clients, respectively. The masking probability in the MLM task is set to 0.15 in the both domain post-training and fine-tuning process. In the fine-turning stage, we initialize weights of feed-forward layers with normal distribution. We set the training epoch as ten and select the model that achieves the best macro-F1 value on the validation set to test on the test set. For both training processes, we adopt cross-entropy loss as the default classification loss. And we use Adam optimizer to train the network with momentum values {$[\beta_1, \beta_2] = [0.9 , 0.999]$}. The learning rate is initialized to $5e-5$ and decayed by using the linear scheduler. The batch size in the training stage is 8. The domain post-training experiment is performed on four NVIDIA A100 GPU, and all the fine-tuning experiments are performed on one NVIDIA A100 GPU. Each fine-tuning experiment costs about 80 minutes.

\subsection{Experimental Results}
\label{experimental_results}
Table~\ref{tab:detailed_experimental_results} shows detailed experimental results about precision, recall and macro-f1 for each category in predicting counselors' intentions and strategies, and clients' reactions and behaviors.

\begin{table}[ht]
    \scalebox{0.6}{
    \centering
    \begin{tabular}{@{}c|c|ccc@{}}
    \toprule
    \textbf{Task} & \textbf{Categories}  & \textbf{Precision} & \textbf{Recall} & \textbf{Macro-f1}   \\ \midrule
    \multirow{3}{*}{Intentions} & Supporting &  0.9194  &  0.8146  &  0.9208 \\
     & Challenging & 0.9578  &  0.6902  &  0.8940 \\
     & Others & 0.9382  &  0.7473  &  0.9072 \\ \hline
     \multirow{13}{*}{Strategies} & Restatement & 0.719  &  0.8891  &  0.795 \\
     &  Reflection of Feelings & 0.7955  &  0.5882  &  0.6763 \\
     & Self-disclosure & 0.5714  &  0.5714  &  0.5714 \\
     & Inquiring Subj. Info. & 0.8447  &  0.8671  &  0.8558 \\
     & Inquiring Obj. Info. & 0.8248  &  0.75  &  0.7856 \\
     & Affirmation \& Reassurance & 0.8055  &  0.76  &  0.7821 \\
     & Minimal Encouragement & 0.9518  &  0.9478  &  0.9498\\
     & Answer & 0.6522  &  0.4286  &  0.5172 \\
     & Interpretation & 0.664  &  0.5773  &  0.6176 \\
     & Confrontation & 0.6667  &  0.2222  &  0.3333\\
     & Invite to Explore New Actions & 0.7717  &  0.7899  &  0.7807 \\
     & Invite to Take New Perspectives & 0.3824  &  0.2766  &  0.321 \\
     & Others & 0.9342  &  0.9148  &  0.9244 \\ \midrule
     \multirow{3}{*}{Reactions} & Positive & 0.9642  &  0.9757  &  0.9699\\
     & Negative  & 0.459  &  0.28  &  0.3478 \\
     & Others & 0.9002  &  0.9043  &  0.9023 \\ \hline
     \multirow{12}{*}{Behaviors} & Giving Information & 0.8952  &  0.9263  &  0.9105 \\
     & Confirming & 0.8881  &  0.9237  &  0.9055 \\
     & Reasonable Request & 0.8468  &  0.8268  &  0.8367 \\
     & Extending & 0.4384  &  0.3546  &  0.3921 \\
     & Reformulating & 0.1  &  0.0345  &  0.0513 \\
     & Expressing Confusion & 0.4545  &  0.4545  &  0.4545 \\
     & Defending & 0.2963  &  0.1569  &  0.2051 \\
     & Self-criticism or Hopelessness & 0.75  &  0.2308  &  0.3529 \\
     & Others & 0.9036  &  0.918  &  0.9107 \\ \hline
    \end{tabular}}
    \caption{The RoBERTa classification result for each category in four tasks, including predicting counselors' intentions, strategies and clients' reactions and behaviors.}
    \label{tab:detailed_experimental_results}
\end{table}

\section{Application to Counseling}
\subsection{Correlation Between Clients' Reactions and Conversation Outcomes}
We group all conversations according to the proportion of the clients' \textit{Negative} reactions contained in the conversations, ensuring that the number of conversations in each group is almost the same (except for the first group). We then calculate the mean and standard deviation of the clients' self-reported conversation-level scores in each group. The results are shown in Table~\ref{tab:negative_reaction_and_outcome}.

\begin{table}[ht]
\centering
\scalebox{0.7}{
\begin{tabular}{@{}cccc@{}}
\toprule
\textbf{Group} & \textbf{Ratio Span} & \textbf{\# Session} & \textbf{Score}  \\
\midrule
1              & 0.000 $\sim$ 0.012    & 181          & $16.62_{3.62}$ \\
2              & 0.012 $\sim$ 0.024    & 37           & $16.76_{3.60}$ \\
3              & 0.024 $\sim$ 0.036    & 47           & $16.81_{3.72}$ \\
4              & 0.036 $\sim$ 0.048    & 39           & $16.33_{3.65}$ \\
5              & 0.048 $\sim$ 0.060    & 35           & $15.74_{3.19}$ \\
6              & 0.060 $\sim$ 0.096    & 42           & $14.74_{4.48}$ \\
7              & 0.096 $\sim$ 0.240    & 38           & $14.16_{5.07}$ \\
\bottomrule
\end{tabular}}
\caption{Grouped conversations according to the ratio of clients' \textit{Negative} reactions included. The last column shows the average scores of conversations in each group, with standard deviations as subscripts.}
\label{tab:negative_reaction_and_outcome}
\end{table}

\subsection{Which behavior influences conversation effectiveness the most?}
\label{important_beavhior}
\begin{table}[ht]
    \centering
    \scalebox{0.28}{
    \begin{tabular}{|c|c|c|c|c|c|c|c|c|c|c|c|}
        \toprule
        \# Variables & Confirming & \makecell[c]{Giving \\ Information} & \makecell[c]{Reasonable \\ Request} & Extending & Reformulating & \makecell[c]{Expressing \\ Confusion} & Defending & \makecell[c]{Self-criticism \\ or \\ Hopelessness} & \makecell[c]{Shifting  \\Topics} & \makecell[c]{Focus \\ Disconnection} & \makecell[c]{Sarcastic \\ Answer} \\ \midrule
        9 &  $\checkmark$ &   $\checkmark$ &  $\checkmark$ &  $\checkmark$ &  $\checkmark$ &  $\checkmark$ &  $\checkmark$ &  $\checkmark$ &  &  &  $\checkmark$\\ \hline
        7  &  $\checkmark$ &  $\checkmark$ &   $\checkmark$ &  $\checkmark$ &  $\checkmark$ &  &  $\checkmark$ &  $\checkmark$ &  &  & \\\hline
         6   &  $\checkmark$ &  $\checkmark$ &  $\checkmark$ &  $\checkmark$ &  $\checkmark$ &   & $\checkmark$ &  &  &  & \\\hline
         5     &  $\checkmark$ &  $\checkmark$ &  &  $\checkmark$ &  $\checkmark$ &  &  $\checkmark$ &  &  &  & \\\hline
         3  &  $\checkmark$ &  $\checkmark$ &  & $\checkmark$  &  &  &  &  &  &  & \\\hline
         2  &  &  $\checkmark$ &  &  $\checkmark$ &  &  &  &  &  &  & \\\hline
         1  &  &  &  &  $\checkmark$ &  &  &  &  &  &  & \\
         \bottomrule
    \end{tabular}}
    \caption{Behavior types selected as important independent variables that affect clients' self-reported evaluation of conversation effectiveness by Lasso model, as the coefficient of L1 regularization uniformly increases from 0.001 to 0.1.}
    \label{tab:lasso_variable_selection}
\end{table}

\subsection{Clients Reactions and Behaviors towards Counselors' Strategies}
\label{clients_behavior_distribution}

\begin{figure*}[ht]
    \centering
    \scalebox{0.22}{\includegraphics{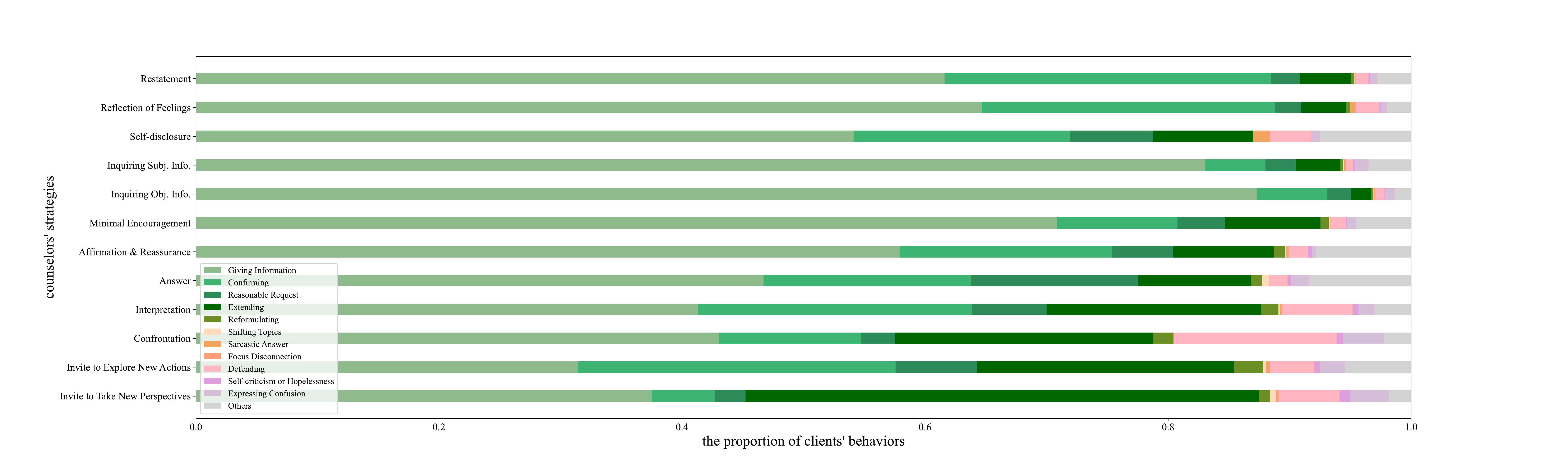}}
    \caption{The clients' behaviors distribution after the counselors' each strategy in the overall conversations.}
    \label{fig:accumulated_strategies_and_followup_behaviors}
\end{figure*}

Figure~\ref{fig:accumulated_strategies_and_followup_behaviors} shows clients' follow-up behavior distribution after the counselor's every strategy in the overall conversations, where the behavior distribution refers to the proportion of the clients' each immediate behavior type. We find that compared with using strategies with \textit{Supporting} intention, counselors' utilization of \textit{Challenging} strategies is more likely to lead to clients' \textit{Negative} behaviors.

We then measure the similarity of the impact of counselors' each strategy on clients' behaviors by calculating the Euclidean distance between clients' follow-up behavior distribution after different strategies (see Table~\ref{tab:distribution_distance}).

\begin{table*}[ht]
\scalebox{0.35}{
\begin{tabular}{lcccccccc|c|cccc|c|}
                                                         & \textbf{Restatement} & \textbf{\makecell{Reflection \\ of Feelings}} & \textbf{Self-disclosure} & \textbf{\makecell{Inquiring \\ Subj. Info.}} & \textbf{\makecell{Inquiring \\ Obj. Info.}} & \textbf{\makecell{Minimal \\ Encouragement}} & \textbf{\makecell{Affirmation \\ \& Reassurance}} & \textbf{Answer} & \textit{\textbf{Supporting}} & \textbf{Interpretation} & \textbf{Confrontation} & \textbf{\makecell{Invite to Explore\\ New Actions}} & \textbf{\makecell{Invite to Take \\ New Perspectives}} & \textit{\textbf{Challenging}}  \\
\textbf{Restatement}  & 0.0000 & 0.0452 & 0.1690 & 0.3069 & 0.3337 & 0.1994 & 0.1359 & 0.2381 & \textbf{0.204} & 0.2691 & 0.3854 & 0.3626 & 0.5131 & 0.3825                  \\
\textbf{Reflection of Feelings}   & 0.0452 & 0.0000 & 0.1724 & 0.2665 & 0.2928 & 0.1648 & 0.1354 & 0.2518 & \textbf{0.1898} & 0.2887 & 0.3854 & 0.3896 & 0.5184 & 0.3955                                                  \\
\textbf{Self-disclosure}  & 0.1690 & 0.1724 & 0.0000 & 0.3369 & 0.3833 & 0.2043 & 0.0630 & 0.1076 & \textbf{0.2052} & 0.1777 & 0.2585 & 0.2767 & 0.4097 & 0.2807                                                        \\
\textbf{Inquiring Subjective Information} & 0.3069 & 0.2665 & 0.3369 & 0.0000 & 0.0582 & 0.1398 & 0.2932 & 0.4122 & \textbf{0.2591} & 0.4847 & 0.5057 & 0.5930 & 0.6073 & 0.5477                                                        \\
\textbf{Inquiring Objective Information} & 0.3337 & 0.2928 & 0.3833 & 0.0582 & 0.0000 & 0.1877 & 0.3388 & 0.4593 & \textbf{0.2934} & 0.5285 & 0.5532 & 0.6380 & 0.6551 & 0.5937                                                          \\
\textbf{Minimal Encouragement}    & 0.1994 & 0.1648 & 0.2043 & 0.1398 & 0.1877 & 0.0000 & 0.1585 & 0.2800 & \textbf{0.1907} & 0.3472 & 0.3882 & 0.4540 & 0.4926 & 0.4205                                                 \\
\textbf{Affirmation and Reassurance}  & 0.1359 & 0.1354 & 0.0630 & 0.2932 & 0.3388 & 0.1585 & 0.0000 & 0.1449 & \textbf{0.1814} & 0.2148 & 0.3020 & 0.3130 & 0.4289 & 0.3147                                                \\
\textbf{Answer} & 0.2381 & 0.2518 & 0.1076 & 0.4122 & 0.4593 & 0.2800 & 0.1449 & 0.0000 & 0.2706 & 0.1591 & 0.2688 & 0.2303 & 0.3917 & \textbf{0.2625}                                          \\ \midrule
\textbf{Interpretation}  & 0.2691 & 0.2887 & 0.1777 & 0.4847 & 0.5285 & 0.3472 & 0.2148 & 0.1591 & 0.3087 & 0.0000 & 0.1921 & 0.1175 & 0.3071 & \textbf{0.2056}             \\
\textbf{Confrontation}    & 0.3854 & 0.3854 & 0.2585 & 0.5057 & 0.5532 & 0.3882 & 0.3020 & 0.2688 & 0.3809 & 0.1921 & 0.0000 & 0.2512 & 0.2661 & \textbf{0.2365}                                            \\
\textbf{Invite to Explore New Actions}  & 0.3626 & 0.3896 & 0.2767 & 0.5930 & 0.6380 & 0.4540 & 0.3130 & 0.2303 & 0.4071 & 0.1175 & 0.2512 & 0.0000 & 0.3107 & \textbf{0.2265}            \\
\textbf{Invite to Take New Perspectives} & 0.5131 & 0.5184 & 0.4097 & 0.6073 & 0.6551 & 0.4926 & 0.4289 & 0.3917 & 0.5021 & 0.3071 & 0.2661 & 0.3107 & 0.0000 & \textbf{0.2946}                                      
\end{tabular}}
\caption{The Euclidean distance between each strategy's follow-up behavior distribution. The columns of \textit{Supporting} and \textit{Challenging} show the average distance of the follow-up behavior distribution between each strategy and all the other strategies belonging to the corresponding intention, where the lower average distance values are bolded.}
\label{tab:distribution_distance}
\end{table*}

\end{document}